\documentclass[letterpaper, 10 pt, conference]{ieeeconf}  

\IEEEoverridecommandlockouts                              

\usepackage{graphicx}
\usepackage{subcaption}
\usepackage{xcolor}
\usepackage{booktabs}       
\usepackage{hyperref}
\hypersetup{
    colorlinks=true,
    linkcolor=blue,
    filecolor=magenta,      
    citecolor=blue,
    urlcolor=blue,
}
\usepackage{algpseudocode}
\usepackage{algorithm}
\usepackage[algo2e]{algorithm2e}
\usepackage{multirow}
\usepackage[normalem]{ulem}
\usepackage{svg}
\usepackage{lipsum}
\usepackage{colortbl} 
\usepackage{amssymb}
\usepackage{pifont}

\usepackage{amsfonts}       
\usepackage{nicefrac}       
\usepackage{tikz}

\usepackage{amsmath,amsfonts,bm}
\usepackage{caption}
\usepackage{subcaption}

\def\eqref#1{equation~\ref{#1}}

\def\1{\bm{1}}

\DeclareMathAlphabet{\mathsfit}{\encodingdefault}{\sfdefault}{m}{sl}
\SetMathAlphabet{\mathsfit}{bold}{\encodingdefault}{\sfdefault}{bx}{n}

\newcommand{\notes}[1]{}

\newcommand{\xxnote}[3]{}
\ifx\hidenotes\undefined
  \renewcommand{\xxnote}[3]{\color{#2}{#1: #3}}
\fi

\newcommand{\method}{AnySkin}
\newcommand{\website}{\url{https://any-skin.github.io/}}

\usepackage[font={small}]{caption}
\renewcommand\thefigure{\arabic{figure}}
\title{\LARGE \bf AnySkin: Plug-and-play Skin Sensing for Robotic Touch}

\author{Raunaq Bhirangi$^{1,2, \dagger}$, Venkatesh Pattabiraman$^{1}$, Enes Erciyes$^{1}$, Yifeng Cao$^{3}$, Tess Hellebrekers$^{4}$, Lerrel Pinto$^{1}$ \\[12pt] 
$^{1}$ New York University \quad $^{2}$ Carnegie Mellon University \quad $^{3}$ Columbia University \quad $^{4}$ Meta AI Research
\thanks{$^{\dagger}$ Correspondence to \texttt{raunaqbhirangi@nyu.edu}} \\[12pt]
\url{https://any-skin.github.io } \\[12pt]
}

\begin{document}

\makeatletter
\let\@oldmaketitle\@maketitle%
\renewcommand{\@maketitle}{\@oldmaketitle%
    \centering
    \includegraphics[width=\linewidth]{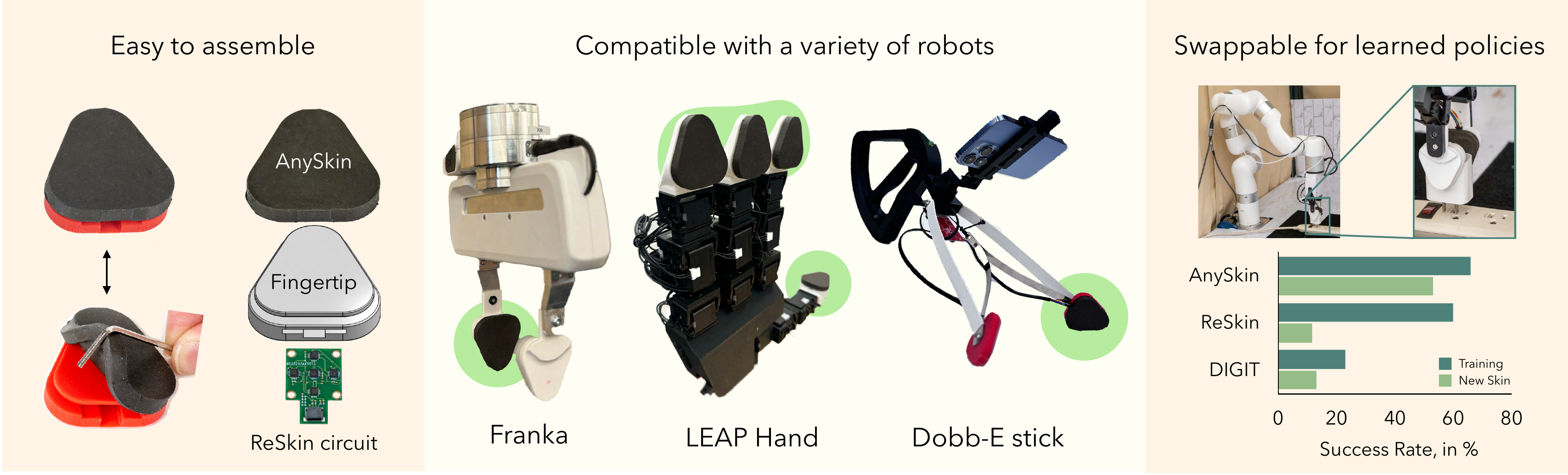}
    \captionof{figure}{We present \method{}, a skin sensor made for robotic touch that is easy to assemble, compatible with different robot end-effectors and generalizes to new skin instances. \method{} senses contact through distortions in magnetic field generated by magnetized iron particles in the sensing surface. The flexible surface is physically separated from its electronics, which allows for easy replacability when damaged.}
    \label{fig:figure1}
}
\makeatother

\maketitle
\thispagestyle{empty}
\pagestyle{empty}


\begin{abstract}
While tactile sensing is widely accepted as an important and useful sensing modality, its use pales in comparison to other sensory modalities like vision and proprioception. \method{} addresses the critical challenges that impede the use of tactile sensing -- versatility, replaceability, and data reusability. Building on the simplistic design of ReSkin, and decoupling the sensing electronics from the sensing interface, \method{} simplifies integration making it as straightforward as putting on a phone case and connecting a charger. Furthermore, \method{} is the first uncalibrated tactile-sensor to report cross-instance generalizability of learned manipulation policies. To summarize, this work makes three key contributions: first, we introduce a streamlined fabrication process and a design tool for creating an adhesive-free, durable and easily replaceable magnetic tactile sensor; second, we characterize slip detection and policy learning with the \method{} sensor; third, we demonstrate zero-shot generalization of models trained on one instance of \method{} to new instances, and compare it with popular existing tactile solutions like DIGIT and ReSkin.
\end{abstract}

\setcounter{figure}{1}

\section{Introduction}

\label{sec:introduction}

Touch sensing is widely recognized as a crucial modality for biological movement and control~\cite{jenner1982cutaneous, johansson1996sensory}. Unlike vision, sound, or proprioception, touch provides sensing at the point of contact, allowing agents to perceive and reason about forces and pressure. However, a closer examination of robotics literature reveals a different narrative. Prominent works and current state-of-the-art in robot learning primarily utilize vision sensing in conjunction with proprioception to train manipulation skills~\cite{Fu_2022_CVPR,padalkar2023open, chi2023diffusion, bharadhwaj2023roboagent}, often ignoring touch. If touch is indeed vital from a biological perspective, why does it remain a second-class citizen in sensorimotor control?

To address this question, let's examine what made cameras ubiquitous in robotics. Three key factors are at play: cost, convenience, and consistency. Cameras are relatively inexpensive (under \$20), easy to integrate on a wide variety of robot platforms (e.g. multi-view, depth, ego-centric), and allow for models trained on them (e.g. object detection, segmentation) to easily transfer to images captured with new cameras. In contrast, touch sensors are often costly due to expensive fabrication processes~\cite{sundaram2019learning} (e.g., pressure-based sensors) or the need for high-end components~\cite{wettels2008biomimetic} (e.g., Biotac). They are inconvenient to use on different robot platforms, being custom-built for specific robot end-effectors and constrained form factors requiring extensive adaptation for different shapes~\cite{wang2021gelsight, taylor2022gelslim}. Finally, touch sensors are inconsistent. Due to boutique fabrication, sensor profiles can vary significantly even when produced through the same process~\cite{bhirangi2021reskin, suresh2023neural}. This inconsistency poses a challenge when transferring tactile-based models across different instances of the same sensor. This transfer is particularly critical for touch sensors due to their persistent need for replacement. Soft sensing interfaces, which are important for touch sensors to maintain a stable grip, wear out more quickly than hard interfaces, resulting in more frequent replacements.

In this work we present \method{}, a new touch sensor that is cheap, convenient to use and has consistent response across different sensor instances. \method{} builds on ReSkin~\cite{bhirangi2021reskin}, a magnetic-field based touch sensor, by improving its fabrication, separating the sensing mechanism from the interaction surface, and developing a new self-adhering, self-aligning attachment mechanism. This allows \method{} to (a) have stronger magnetic fields, which significantly improves its sensor response, (b) be easy to fabricate for arbitrary surface shapes, which allows easy use on different end-effectors, (c) be easy to replace the sensor without adversely affecting the data collection process or the efficacy of models trained on previous sensors (Fig.~\ref{fig:figure1}). 

We run a suite of experiments to understand the efficacy of \method{} vis-a-viz other prominent touch sensors. Our main findings can be summarized below:
\begin{enumerate}
    \item \method{} can readily be used on a variety of robots including xArm, Franka, and the four-fingered Leap hand (See fabrication details in Section~\ref{sec:fabrication}).
    \item \method{} is compatible with ML techniques for slip detection and visuo-tactile policy learning for precise tasks such as inserting USBs (See learning details in Section~\ref{sec:experiments}).
    \item \method{} takes an average of 12 seconds to replace and can be reused after replacement (See replacement study in Section~\ref{subsec:ease-replaceability}).
    \item Models trained on one \method{} transfer zero-shot to a different \method{} with only a 13\% reduction in performance on a plug insertion task compared to the 43\% drop in performance with ReSkin~\cite{bhirangi2021reskin} sensors.
\end{enumerate}

\method{} is fully open-sourced. Videos of fabrication, attachment, and robot policies are best viewed on our project website: \website{}.

\section{Related Work}
\label{sec:related_work}
\subsection{Tactile sensing}
Existing literature on tactile sensing explores a wide range of modalities, each with their own set of advantages and limitations. Capacitative sensors~\cite{glauser2019deformation, wu2020capacitivo, Xu2015StretchNF, Xu2024CushSenseSS} sense contact through changes in capacitance, offering high sensitivity. However, they struggle to model shear force and are prone to breakage due to direct electrical connections between the circuitry and elastomer. Resistive sensors~\cite{sundaram2019learning, bhattacharjee2013tactile, stassi2014flexible} are simple and durable, but tend to provide spatially discrete sensing with low spatial resolution. MEMS-based sensors~\cite{wettels2008biomimetic} offer significant versatility by combining multiple sensors like audio and IMU sensors for multimodal feedback in a mm-scale form factor, but tend to use high-end components and are expensive to fabricate. Optical sensors~\cite{taylor2022gelslim, yuan2017gelsight, lambeta2020digit, di2024usingfiberopticbundles} capture high-resolution contact information using cameras to track the deformation of an elastomer, but often pose hard, stringent limits on the sensor form factor, due to the physical constraints on the camera field of view. This complicates integration for a wide range of applications and significantly increases the effort required to sensorize surfaces of different shapes and sizes.

Magnetic tactile sensors~\cite{hellebrekers2019soft, tomo2018new, Yan2021SoftMS} largely overcome these limitations due to three salient advantages: (a) separation of the sensing electronics from the sensing interface to improve robustness (b) compatibility with different form factors, and (c) an ability to capture shear forces~\cite{bhirangi2021reskin}. Two prominent classes of magnetic sensors in robotics right now - ReSkin~\cite{bhirangi2021reskin} and uSkin~\cite{tomo2018new} - use elastomeric sensing interfaces with magnetic microparticles and macro-sized magnets respectively. In this work, we build on ReSkin sensors due to their lower cost and ease of fabrication.

\subsection{Replaceability for Tactile Sensors}
Recent developments in rapid prototyping and elastomer technology have spurred a substantial rise in the number of robotic tactile sensors. Most tactile solutions rely on soft sensing interfaces to enable stable conformal contact with objects in the environment. Soft interfaces are prone to frequent wear and tear from contact-rich interactions, but discussions on replaceability for tactile sensors remain few and far between. There are two main factors to consider when evaluating replaceability: (a) the physical ease of replacing the sensory interface, and (b) signal consistency when replacing a worn out instance with a new instance. The former is far more frequently discussed~\cite{wettels2008biomimetic, bhirangi2021reskin, lambeta2020digit, gelsightmini2023} and resolved by simply separating the sensing interface -- generally the damage-prone soft elastomer -- from the sensing electronics which last much longer. The latter, however, is much less discussed. Prior work in tactile sensing has found significant variation across different instances of the same tactile sensor~\cite{bhirangi2021reskin, suresh2023neural}. Higher susceptibility to wear combined with lower signal consistency across sensor instances severely restricts the scale of data explored in most existing work on tactile learning~\cite{bhirangi2023all, funabashi2019morphology}. 

This effect is even more pronounced for policy learning where imitation learning as well as reinforcement learning approaches have been used to show impressive results on real-world robots~\cite{padalkar2023open, akkaya2019solving, etukuru2024robot}. However, both approaches rely on significant quantities of data, be it demonstration or online interaction. The necessity of using a single sensor instance across training and testing has severely limited the extent of capabilities demonstrated with visuotactile learning~\cite{lisee, chen2022visuotactile}. Recent research has either relied on policy learning in simulation using simplified models of the tactile sensor~\cite{qi2023general, yin2024learning, yuan2024robot}, or used analytical models for manual feature extraction and dimensionality reduction~\cite{li2014localization, kim2022active}. The former results in a significant dilution in the amount of tactile information and is therefore restricted to less precise tasks with simpler contact reasoning. The latter techniques are often specific to the task they solve, difficult to scale and show limited generalizability beyond the restrictive settings they operate in. In this light, signal consistency across instances is central to building scalable and generalizable tactile models, and making tactile sensing a ubiquitous presence in robot learning. In Section~\ref{sec:experiments}, we quantitatively demonstrate the improved consistency of \method{} signal over ReSkin, and present a direct replaceability comparison with DIGIT \cite{lambeta2020digit} and ReSkin through policy learning.

\begin{figure*}[tbp]
    \begin{subfigure}[b]{0.65\textwidth}
        \centering
        \includegraphics[width=\linewidth]{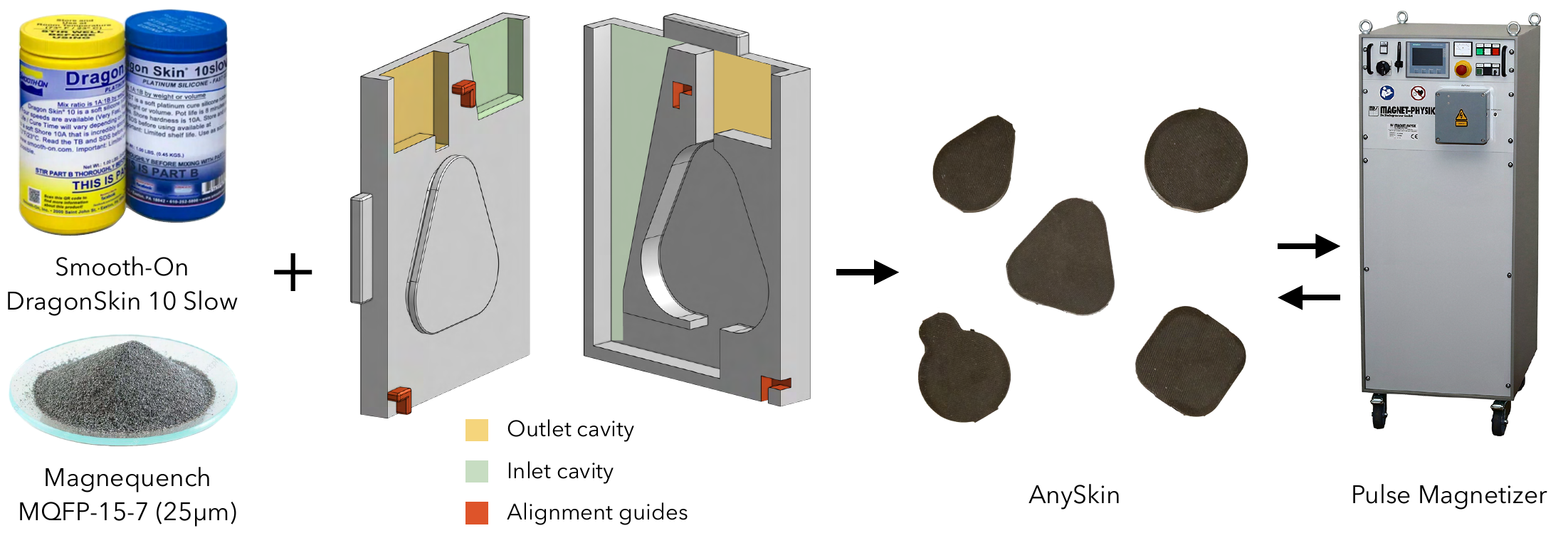}
        \caption{Fabrication of AnySkin}
        \label{fig:anyskin-molds}
    \end{subfigure}
    \hfill
    \begin{subfigure}[b]{0.3\textwidth}
        \centering
        \begin{subfigure}[b]{\linewidth}
            \centering   
            \includegraphics[width=.7\linewidth]{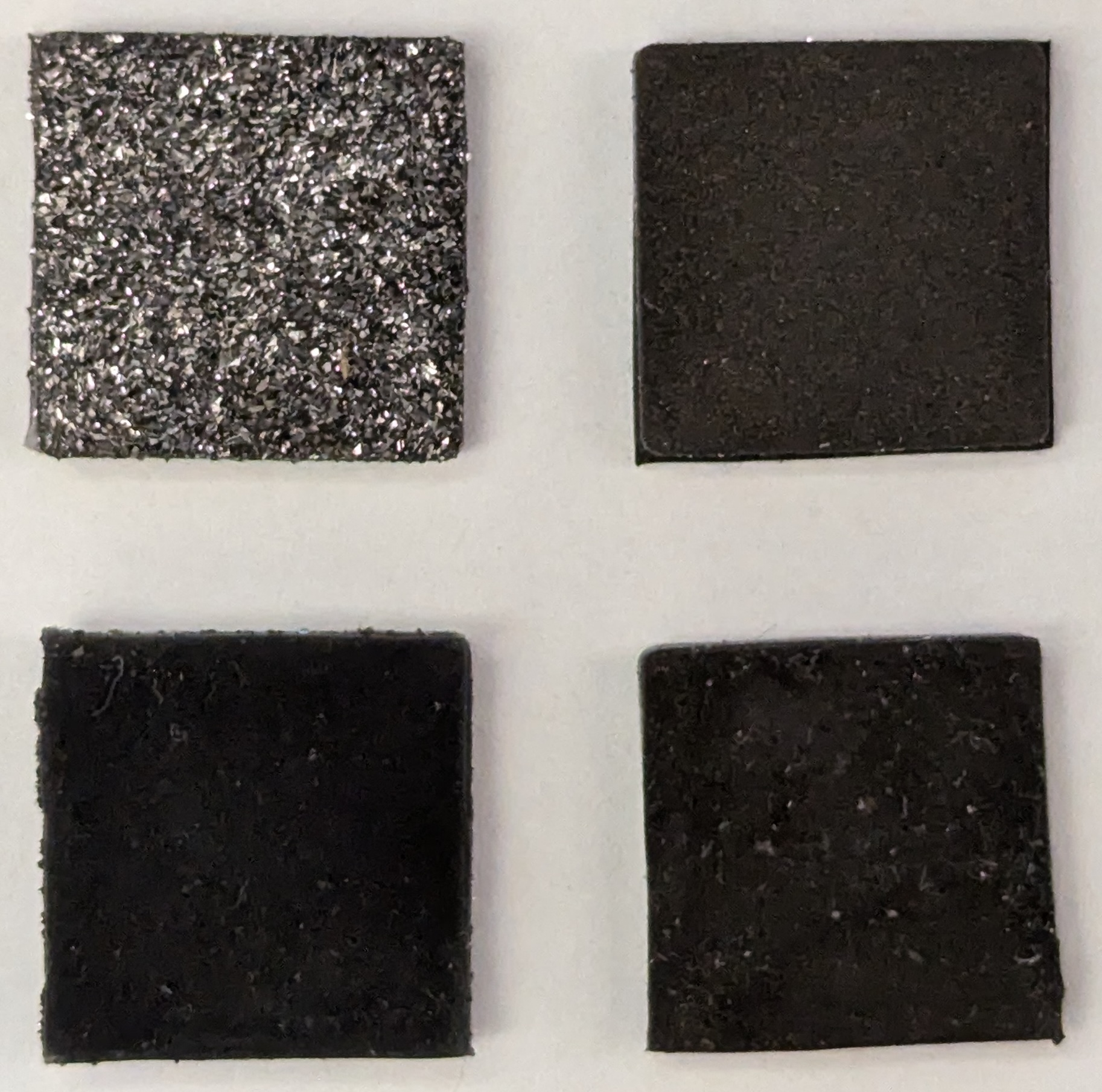}
            \caption{Top and bottom surfaces of skins}
            \label{fig:particle-size-top-bottom}
        \end{subfigure}
    \end{subfigure}
    \caption{(a) AnySkin is made by mixing Smooth-On DragonSkin 10 Slow and MQFP-15-7(25$\mu$m) magnetic particles in a 1:1:2 ratio, and curing it in the two-part molds shown above. Cured skins are magnetized using a pulse magnetizer. (b) Skins made with MQP-15-7(-80 mesh) and MQFP-15-7(25$\mu$m) particles. Note the concentration of particles at the surface of the former due to the larger particle size.}
    \label{fig:fabrication}
\end{figure*}

\section{\method{}: Components}
\label{sec:background}
\method{} builds on ReSkin~\cite{bhirangi2021reskin}, a tactile skin composed of a soft magnetized skin coupled with magnetometer-based sensing circuitry. By detecting distortions in magnetic fields, ReSkin measures skin deformations caused by normal and shear forces~\cite{hellebrekers2019soft, bhirangi2021reskin}. Its adaptability enables integration across various applications, from robotic hands~\cite{bhirangi2023all} to arm sleeves and even dog shoes. \method{} uses the same 5-magnetometer circuitry as ReSkin, while introducing key design and fabrication changes to the skin to improve durability, repeatability, and replaceability.

\begin{itemize}
    \item Magnetizing skins post-curing using a pulse magnetizer.
    \item Introducing physical separation between magnetic elastomer and magnetometer circuit.
    \item Utilizing finer magnetic particles to achieve a more uniform particle distribution.
    \item Implementing a self-aligning design for reduced variability in the positioning of elastomers and circuitry.
\end{itemize}
While some of these additions have been used in isolation in prior work~\cite{bhirangi2023all, sundaram2023dragonclaw, bhirangi2024hierarchical}, there has been little discussion on their effect on sensor response. In this section, we elaborate on the rationale for each choice, followed by a quantitative comparison of the sensor response in Section~\ref{subsec:reskin-v-anyskin}.


ReSkin uses a grid of cube magnets during curing to magnetize the elastomeric skins. While effective, this approach has several drawbacks, such as producing skins with relatively weak magnetic fields. As a result, although the design of ReSkin separates the sensing electronics and the sensing interface, adding physical distance between the skin and the sensors significantly weakens the signal, making it infeasible in reality. In contact-rich tasks, where the sensing skin undergoes considerable strain, the absence of physical separation leads to stress being transmitted directly to the electronics, ultimately compromising their durability.

Additionally, applying a magnetic field during elastomer curing increases variability in the signal response. Before curing, magnetic particles are free to move through the liquid elastomer under the effect of the magnetic field. As a result, the distribution of particles is influenced by the temporal evolution of the applied magnetic field, i.e. how you move the magnets into place, which can be difficult to control when fabricating. To circumvent these disadvantages, we propose using a pulse magnetizer to magnetize the skins post-curing in line with ~\cite{bhirangi2023all, bhirangi2024hierarchical}, as shown in Fig.~\ref{fig:anyskin-molds}. The pulse magnetizer can apply a large enough magnetic field to magnetize the dipoles in the magnetic elastomer. Curing outside the influence of magnetic fields allows for a more uniform distribution of magnetic particles through the bulk of the sensor, thereby improving magnetic field consistency.

However, simply changing the magnetization procedure results in other problems. Curing outside the influence of a magnetic field causes the particles to settle to the bottom of the sensing skin due to gravity as shown in Fig.~\ref{fig:particle-size-top-bottom}. This results in lower durability as the skin begins to shed magnetic particles, particularly during contact-rich interactions. To get around this problem, we replace the magnetic particles with much finer particles (details in Section~\ref{subsec:elastomer-fab}). The smaller particles operate in a sufficiently low Reynolds number regime to allow the elastomer to cure before they can settle on one surface of the elastomer. 

Finally, since ReSkin relies on magnetic field distortions to measure contact characteristics, sensor response is strongly influenced by the relative positioning of the magnetic skin and the magnetometer circuitry (see Section~\ref{subsec:reskin-v-anyskin}). This means that loss of adhesion, peeling, or any other relative motion between skin and circuitry over the life of the sensing skin can significantly affect the consistency of the signal. Ideally, we would like the skin to stay adhered until it needs to be replaced due to wear and tear. Using screws to adhere the skin as suggested in~\cite{bhirangi2021reskin} results in poor durability due to a stress concentration at the screw-skin interface, especially in tasks involving shear forces. Using an adhesive like Silpoxy~\cite{lambeta2020digit, bhirangi2023all} can be used to create sticker-like skins that last relatively longer but still tend to peel under repetitive cyclic loading. With \method{}, we eliminate the need for adhesives or fasteners by modifying the design of the skins to be self-adhering. Additionally, we also eliminate the need to manually align skin and circuitry, significantly improving signal consistency as demonstrated in Section~\ref{subsec:reskin-v-anyskin}.

\begin{table*}[!htbp]
    \centering
    \caption{AnySkin's signal strength is comparable to ReSkin with lower variability across instances, and physical separation from the magnetometer electronics. Statistics computed over 5 samples of each type (PM: Pulse magnetizer, FP: finer particles, SA: self-aligning).}
    \begin{tabular}{lcccccccc}
        \multirow{4}{*}{Experiment} & \multicolumn{2}{c}{\includegraphics[height=1cm]{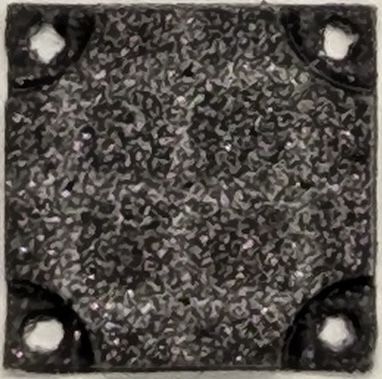}} & \multicolumn{2}{c}{\includegraphics[height=1cm]{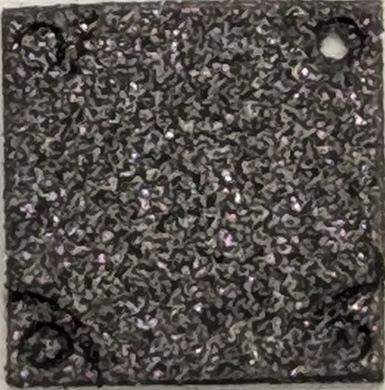}} & \multicolumn{2}{c}{\includegraphics[height=1cm]{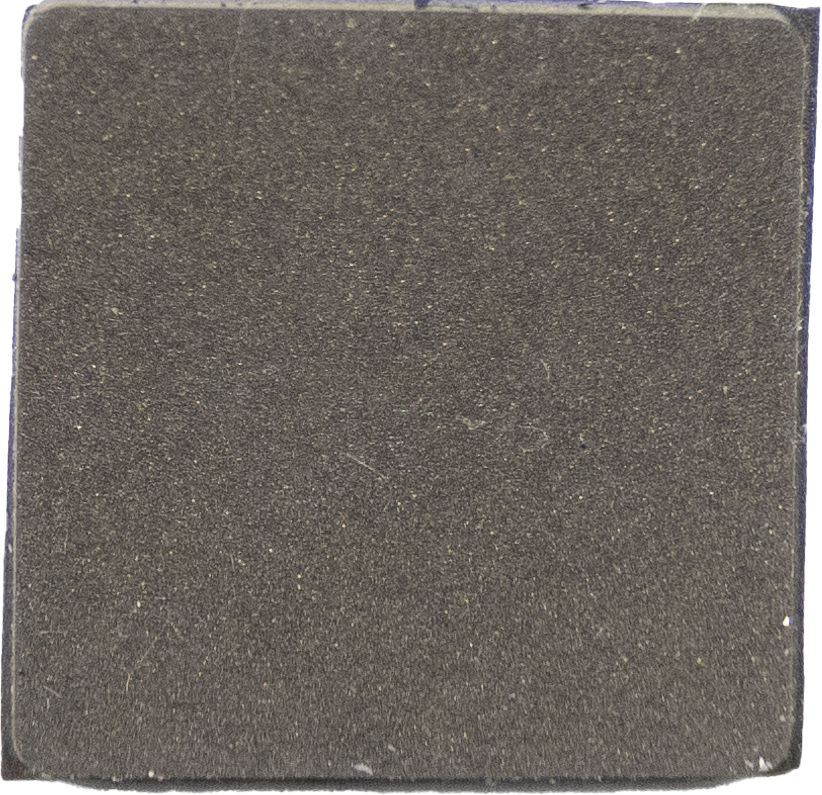}} & \multicolumn{2}{c}{\includegraphics[height=1.5cm]{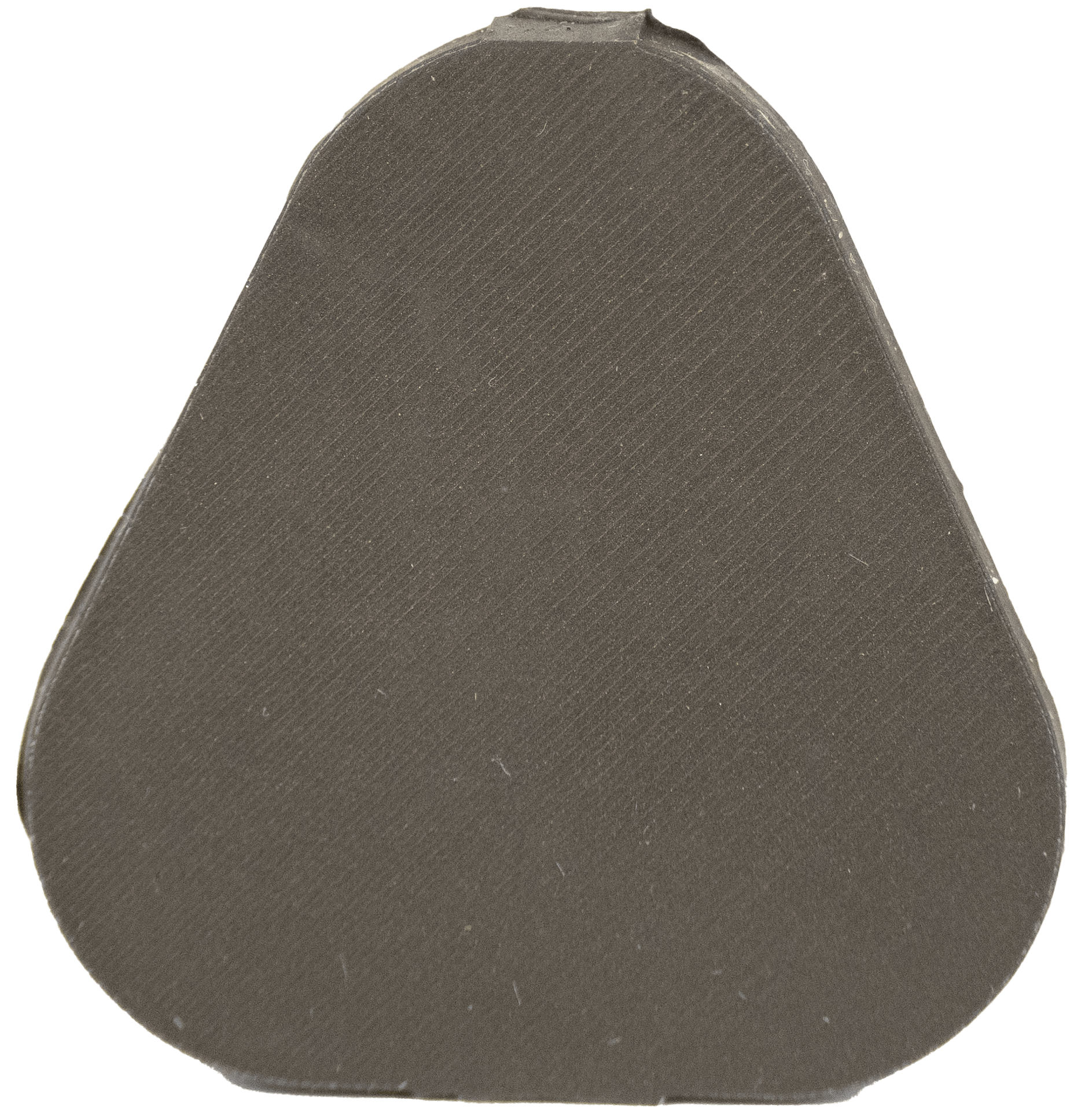}} \\
          & \multicolumn{2}{c}{ReSkin} & \multicolumn{2}{c}{\textcolor{green}{+PM}} & \multicolumn{2}{c}{\textcolor{green}{+FP}} & \multicolumn{2}{c}{\textcolor{green}{+SA}  (AnySkin)} \\
        
         & $B_{xy}$ & $B_z$ & $B_{xy}$ & $B_z$ & $B_{xy}$ & $B_z$ & $B_{xy}$ & $B_z$ \\\midrule
         Average strength, in $\mu$T & 1062 & 302 & \textbf{1818} & \textbf{5212} & \textbf{1602} & \textbf{5784} & 283 & 1265 \\
         Normalized std. deviation across instances & 0.54 & 0.87 & 0.34 & 0.12 & 0.21 & 0.15 & \textbf{0.12} & \textbf{0.10} \\
         Normalized std. deviation across 1mm misalignments & 1.38 & 1.43 & 0.25 & 0.11 & 0.18 & 0.07 & \multicolumn{2}{c}{\textbf{Self-aligning}} \\\bottomrule
         
         \bottomrule
    \end{tabular}
    \label{tab:signal-comparison}
\end{table*}

\section{\method{}: Fabrication}
\label{sec:fabrication}
The overall fabrication procedure follows the general outline of ReSkin: Magnetic particles and elastomer are mixed in specified ratios; the resulting mixture is poured into the molds; cured skins are magnetized. The shape of the fingertip-skin assembly is designed to be triangular as shown in Fig.~\ref{fig:figure1} to improve reachability. In this section, we elaborate on the details of the fabrication procedure for \method{}, and key changes to the ReSkin fabrication procedure that result in a new, upgraded sensor.

\subsection{Mold design}
The shape of the magnetic skin is dictated by the molds used for curing. To create self-adhering skins as outlined in Section~\ref{sec:background}, we present a two-part mold design as shown in Fig. \ref{fig:anyskin-molds}. We choose a skin thickness of 2 mm following \cite{bhirangi2021reskin} with a triangular shape for its advantageous form factor for precise manipulation. All the experiments presented in this paper use this triangular skin. We also open-source a mold design CAD tool that generates designs for the fingertip as well as 2-part molds from just a 2D drawing. Unlike tactile sensors that require significant engineering for changes in form factor~\cite{lambeta2020digit, taylor2022gelslim}, \method{} makes it effortless to diversify your tactile sensor.

\subsection{Elastomer composition}
\label{subsec:elastomer-fab}
For \method{}, we mix magnetic microparticles and two-part polymer (Dragonskin 10 Slow; Smooth-On) in the same 2:1:1 ratio as ReSkin, while using finer Magnequench MQFP-15-7(25$\mu$m). These particles are about 100x smaller than the MQP-15-7(-80 mesh) used with ReSkin, and do not settle before curing, due to their lower Reynolds number~\cite{Falkovich_2011}. This ensures that magnetic particles are more evenly distributed through the volume of the skin, thereby improving consistency of the signal.

\subsection{Magnetization}
ReSkin is magnetized by sandwiching the magnetic elastomer mix between a grid of magnets while it is curing. This results in higher variance in distribution of magnetic particles within the core of the skin based on the exact timing of sandwiching the skins. Drawing from D'Manus~\cite{bhirangi2023all}, we use a pulse magnetizer for magnetizing the skins after curing is complete. Separating the magnetizing process from the curing process allows the skins to cure undisturbed and maintain a more uniform distribution of magnetic particles. Furthermore, the magnetic field applied by the pulse magnetizer is far stronger than the sandwich of grid magnets. This results in skins with stronger magnetic fields, which in turn enables larger separations between magnetic skin and magnetometer circuitry.

\subsection{Magnetic elastomer fabrication}
The final fabrication process follows similar steps as the ReSkin fabrication process. The molds are first aligned using the built-in alignment guides and clicked together. We use plastic clamps to hold the parts together. The two-part elastomer compound is then mixed and degassed. This is followed by the addition of magnetic micro particles and another round of mixing and degassing. Once degassing is complete, the magnetic elastomer mix is poured through the mold inlet as shown in Fig.~\ref{fig:fabrication} until it emerges at the outlet, pausing as necessary to allow the mixture to flow through and fill the entire mold. The filled mold is then placed in a vacuum chamber and a pressure of 29mm of Hg/in is applied, again pausing as necessary to prevent overflow as the liquid bubbles. This pressure is held for 10 minutes before releasing the vacuum. The molds are allowed to rest for 16 hours, before prying them open and trimming excess material to reveal the fully cured \method{}.

\begin{figure*}[tbp]
    \centering
    \includegraphics[width=\textwidth]{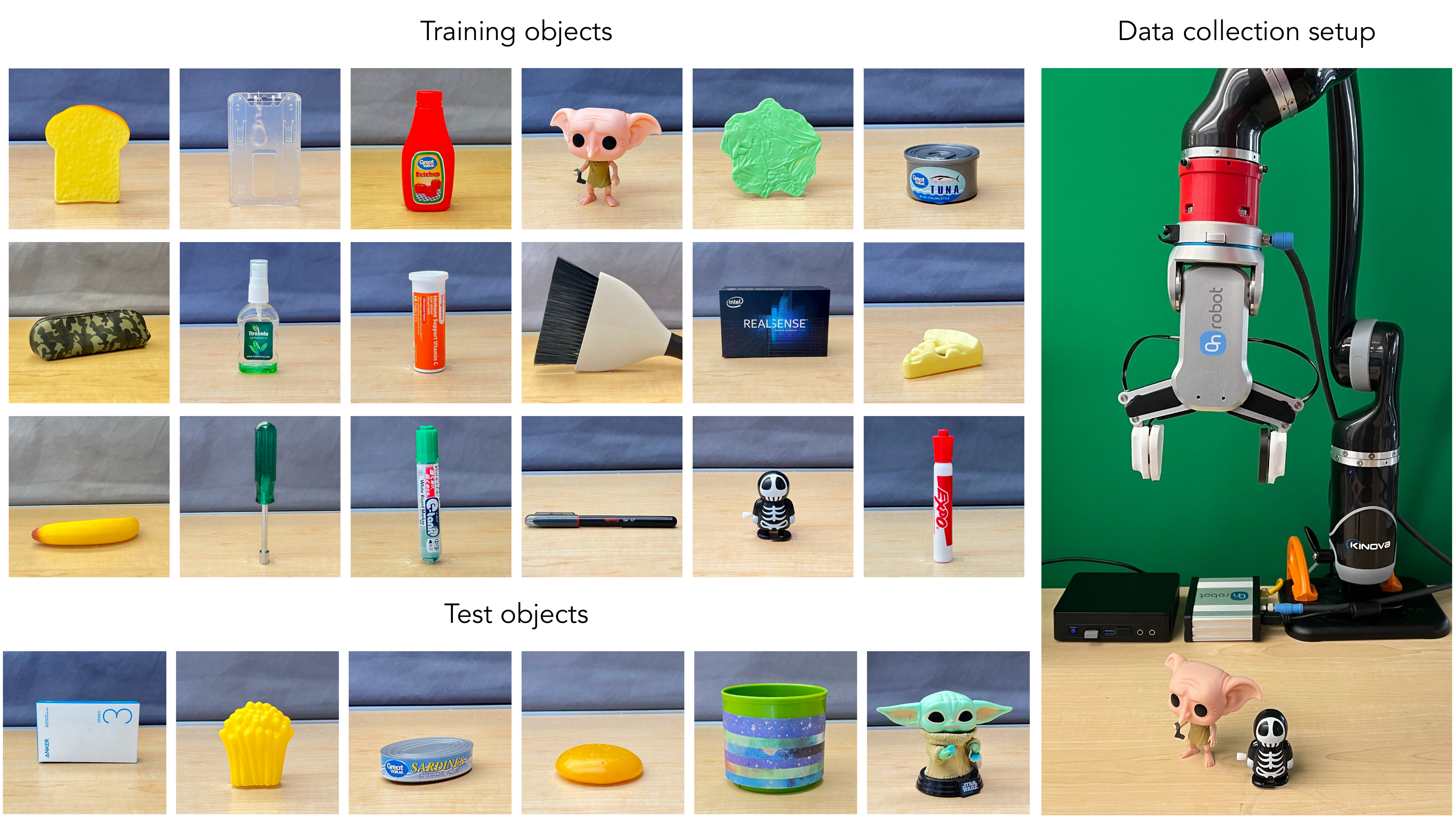}
    \caption{Experimental setup used for slip detection experiments, where we train LSTM models on data collected by a Jaco Robot equipped with the \method{} sensor (right). We train on a set of training objects (left top) and evaluate it on a set of unseen test objects (left bottom).}
    \label{fig:slip-expt-setup}
\end{figure*}

\section{Experiments and Results}
\label{sec:experiments}
In this section, we perform extensive experiments to demonstrate the capabilities of \method{} as a tactile sensor, and within the context of policy learning. These experiments are designed to answer the following questions:
\begin{itemize}
    \item How do the fabrication changes outlined in Section~\ref{sec:background} influence signal characteristics?
    \item Can \method{} sensors be used to detect slip?
    \item How does \method{}'s ease of replaceability compare with other sensors like DIGIT and ReSkin?
    \item How does replacing \method{} affect the performance of learned policies, and compare with other sensors like ReSkin and DIGIT?
\end{itemize}

\subsection{Comparison between ReSkin and \method{} signal}
\label{subsec:reskin-v-anyskin}

To quantitatively demonstrate the effect of each of the fabrication changes listed in Section~\ref{sec:background} towards improving the consistency of \method{}, we present the following set of experiments analyzing the raw signal from the four different skins shown in Table~\ref{tab:signal-comparison}, tracking the progression from ReSkin to \method{}:

\subsubsection{Effect of pulse magnetizer on signal strength}
To understand the effect of the pulse magnetizer on signal strength, we take five instances of each skin type and measure the raw signal corresponding to each instance. We average the absolute values across the three axes of the five magnetometers, and report the results in Table~\ref{tab:signal-comparison}. We see a significant increase in the raw magnetic field for both sets of pulse magnetized skins. This increase allows us to add a physical separation between sensing skin and the sensory electronics, which improves replaceability, as well as repeatability of the signal as discussed below.

\subsubsection{Comparison of signal consistency across skins}
To compare signal consistency across the sensing skins, we compute the standard deviation along each axis of the five magnetometers across the five instance of each skin type. To account for the larger signal strengths of the pulse magnetized skins and allow for a fairer comparison, we normalize the computed standard deviations by the mean absolute values along $xy$ and $z$ axes for each skin type. Aggregated statistics for the different skins are presented in Table~\ref{tab:signal-comparison}. Note that finer particles reduce the variability in sensory signal, which could be attributed to the more uniform distribution of particles resulting from reduced settling under the influence of gravity.

\subsubsection{Effect of alignment on signal consistency}
\label{sec:alignment}
\begin{figure}[htbp]
    \centering
    \includegraphics[width=\linewidth]{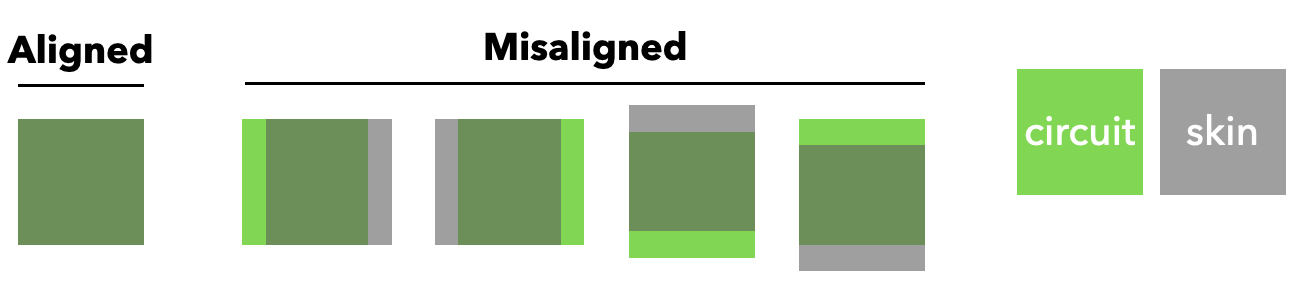}
    \caption{Different circuit-skin alignments evaluated in Section~\ref{sec:alignment}}
    \label{fig:offset-expt}
\end{figure}

Finally, to understand the importance of the self-aligning design of \method{}, we take a single skin (20mm $\times$ 20mm) of each type except \method{}, and collect magnetometer data from placing it at a 1mm offset along each side as shown in Fig.~\ref{fig:offset-expt}. Normalized standard deviations are computed across the aligned and misaligned variants across each skin, and results are reported in Table~\ref{tab:signal-comparison}. Across different skins, variability in sensor response from just misalignment is larger than the cross-instance variability seen with \method{}, underlining the importance of the self-aligning design of \method{}.

\begin{figure*}[tbp]
    \centering
    \includegraphics[width=\textwidth]{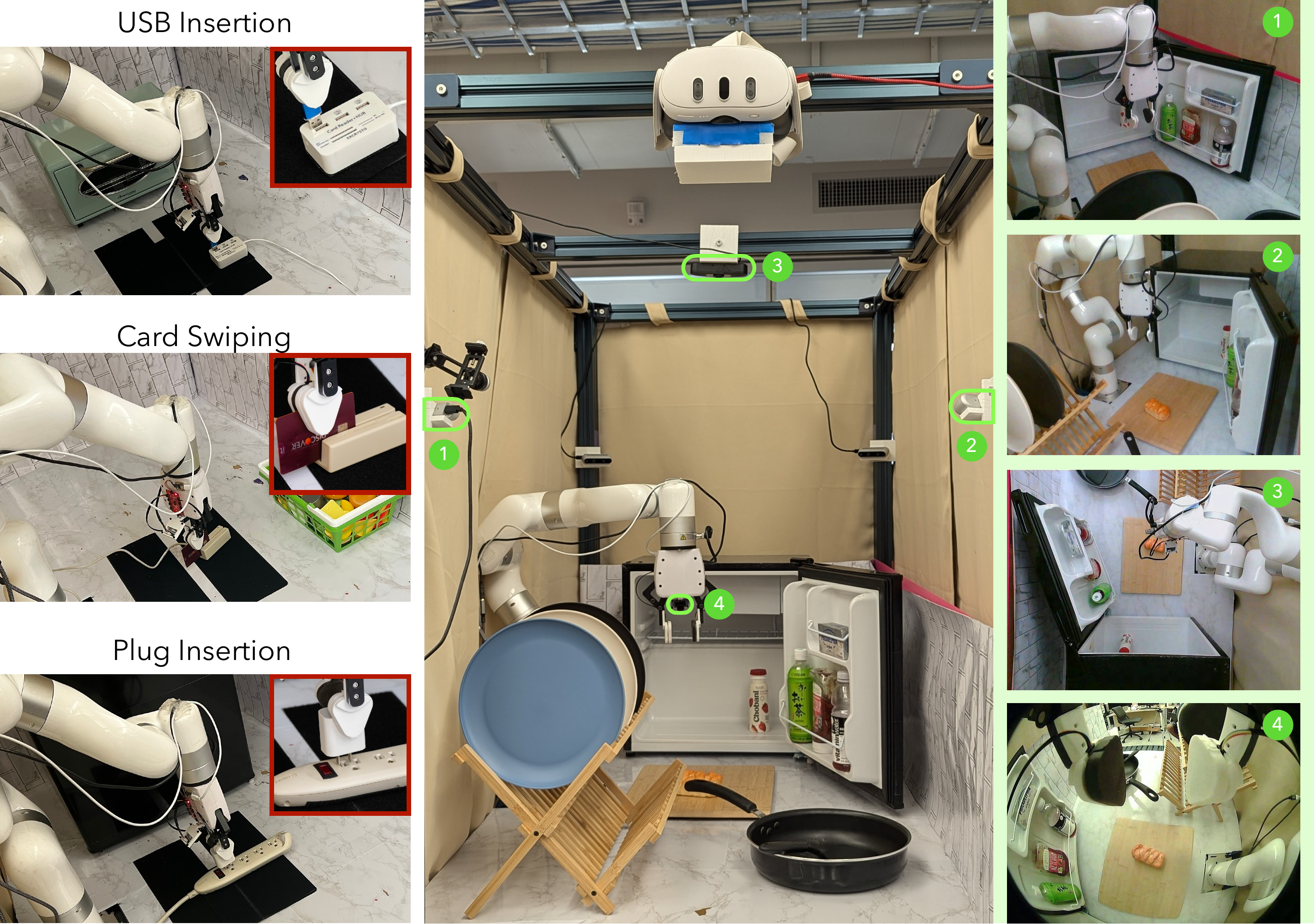}
    \caption{We evaluate the replaceability of \method{} on a set of 3 precision tasks, where capturing the contact interaction using touch is critical (left). Our experimental setup consists of a Ufactory xArm 7 robot with an \method{} sensor integrated into the standard gripper (center). Visual information is captured using three static cameras (1-3) and one egocentric camera (4) attached to the gripper. (right)}
    \label{fig:expt-setup}
\end{figure*}

\subsection{Slip Detection}

We quantify \method{}’s ability to detect slip through a controlled experiment. Our setup consists of a Kinova Jaco arm and an Onrobot RG-2 gripper with integrated \method{}. An object held by a human operator is grasped and lifted up slowly for 1 second. We use a set of 40 daily objects -- 30 for training and 10 evaluation -- with varying shapes, weights and materials. We collect 6 trajectories for each object by changing the grasping force, width and location. After the data collection is complete, a human annotator labels the sequence as slip or no-slip from the corresponding videos. In contrast with \cite{li2018slipdetectioncombinedtactile}, we only use tactile signals as input to the slip detection model. Some of the objects used, along with the data collection setup are shown in Fig. \ref{fig:slip-expt-setup}. The full set of training and test objects as well as videos of the learned policy can be found on our website.

The data collection frequency for tactile data is 100 Hz. We subsample the signal by 15 along the temporal axis and take the first difference. We use an LSTM \cite{hoch1997longshort} to train our slip prediction models. Our model is able to detect slip on unseen objects with 92\% accuracy.

\subsection{Ease of replaceability}
\label{subsec:ease-replaceability}
We compare the ease of replaceability of \method{} against the replaceability of other skins like DIGIT and ReSkin. For ReSkin, we use two different methods of adhering skins to the circuit board -- screws~\cite{bhirangi2021reskin} and adhesive stickers~\cite{bhirangi2023all}. The skins are compared along two axes -- replacement time and re-usability after replacement -- through a user study with 10 non-expert users, and the results are presented in Table~\ref{tab:replaceability}

\begin{table}[htbp]
    \centering
    \caption{Comparison of replaceability of different sensors}
    \begin{tabular}{ccc}
         Sensor & Time to replace, in $s$ & Reusable \\ \midrule
         ReSkin (adhesive) & $82 \pm 64$ & No \\
         ReSkin (screws) & $ 236 \pm 64$ & Yes \\
         DIGIT & $58 \pm 22$ & Yes \\ 
         \method{} & $\mathbf{12 \pm 5}$ & Yes \\\bottomrule
    \end{tabular}
    \label{tab:replaceability}
\end{table}

We find that users find it significantly faster and easier to replace \method{} than any of the other skins used for comparison. Furthermore, eliminating adhesion allows replaced skins to be reused without needing the extra hours to reapply adhesion and allow it to cure.

\subsection{Replaceability in Policy Learning}
The most important consequence of the signal consistency and replaceability of \method{} outlined so far, is its ability to enable policy generalization across different instances of the skin. In this section, we demonstrate the cross-instance generalizability of \method{} across three precise manipulation tasks. We follow this up with a comparison of the cross-instance generalizability of policies trained on DIGIT, ReSkin and \method{} on the plug insertion task.


\begin{figure*}[htbp]
    \centering
    \includegraphics[width=\linewidth]{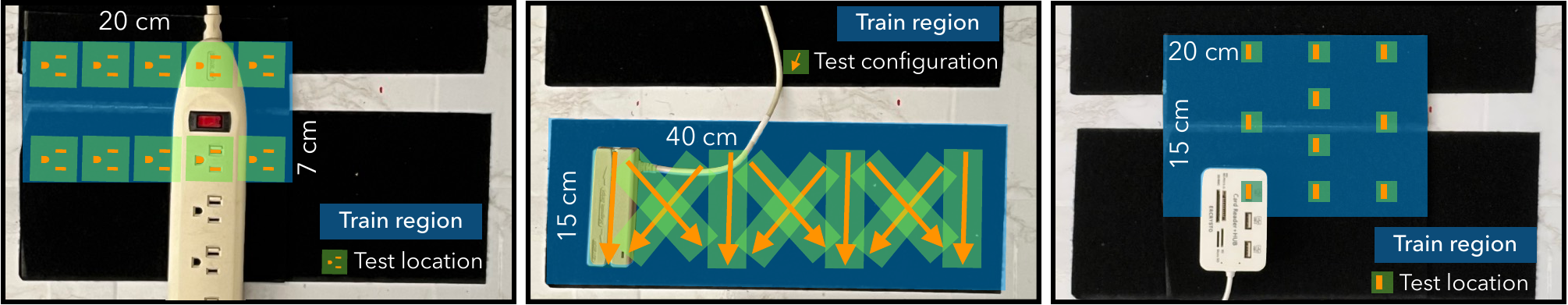}
    \caption{Training and test locations of the target objects interacted with for each task. The blue region represents the extent of variation in the location of the target object, while the green-orange blocks denote held-out test configurations used for evaluation.}
    \label{fig:task-variations}
\end{figure*}

\subsubsection{Experimental Setup}
For our policy learning experiments, we train behavior cloning models for a set of precise manipulation tasks. Our experimental setup is shown in Fig.~\ref{fig:expt-setup} and consists of an X-Arm 7DOF robot in a robot cage as shown in the image. The setup consists of four different cameras -- three fixed to the frame and one egocentric camera attached to the robot's wrist. We attach an \method{} sensor to one of the fingertips and a plain silicone tip to the other. A Meta Quest 3 and the accompanying joystick controller are used to teleoperate the robot using Open-Teach~\cite{iyer2024open}, an open-source teleoperation framework.

We demonstrate the replaceability of \method{} on a set of three contact-rich manipulation tasks shown in Fig.~\ref{fig:expt-setup}:
\begin{itemize}
    \item \textbf{Plug insertion}: The robot starts with a plug grasped in the gripper. The task requires the robot to move to the first socket on the socket strip and insert the plug. The location of the socket strip is randomized in a 20 cm $\times$ 7 cm box on the table, and learned policies are tested on socket locations unseen in the training data.
    \item \textbf{Card swiping}: The robot starts with a credit card grasped in the gripper. It must move to the location of a credit card machine on the table and swipe the credit card. The location of the credit card machine is randomized in a 40 cm $\times$ 15 cm box and angled in the range $(-30^\circ \text{ to } 30^\circ)$ on the table, and learned policies are tested on card machine locations unseen in the training data.
    \item \textbf{USB insertion}: The robot starts with a USB cable grasped in the gripper. It must move to the location of the USB port on the table and insert the cable. The location of the USB port is randomized in a 20 cm $\times$ 15 cm box on the table, and learned policies are tested on port locations unseen in the training data.
\end{itemize}

To ensure that learned policies rely on \textit{both} vision and tactile information, we vary the configuration of the target object, ie. the socket strip, the card machine and the USB port for plug insertion, card swiping and USB insertion respectively in the demonstration dataset. For all the evaluations presented here, we use a set of held-out configurations of the target object as shown in Fig.~\ref{fig:task-variations}.

\subsubsection{Model Architecture and Training}
Our policies are trained using behavior cloning. For each task presented in this section, we collect a set of 96 demonstration trajectories, with data from the four cameras in addition to unchanged instances of the respective tactile sensor(s) being used. The BAKU~\cite{haldar2024baku} architecture is used as the policy architecture. BAKU tokenizes each input using a modality-specific encoder: image inputs from cameras and DIGITs are encoded using ResNet-18~\cite{he2016deep} encoders, while \method{} and ReSkin inputs are encoded using an MLP encoder. An action token is appended to the set of encoded tokens before passing the sequence through a transformer encoder, and the output corresponding to the action token is used to predict actions. We use action chunking~\cite{zhao2023learning} and predict the next 10 actions at every timestep. For every training setting, we train three separate models corresponding to three different seeds, and present aggregated statistics on 10 policy rollouts.

\subsubsection{Evaluating cross-instance generalizability}
\label{subsec:replaceability-policy}
To investigate the replaceability of \method{} in the context of policy learning, we evaluate behavior cloning policies trained using a single instance of \method{} on a new instance. Note that we use a different training and test skin for \textit{each} of the presented tasks to avoid over-indexing on specific skin instances. Table~\ref{tab:skin-swapping} presents a comparison between policy performance with the original and swapped skins for each of the precise, contact-rich tasks outlined above. Additionally, we train and evaluate a policy that only takes camera images as input to serve as a control experiment and verify that the policies indeed rely on tactile data. We find that across tasks, performance drops by an average of just 15.6\% and visuotactile policies with swapped skins continue to do significantly better than purely visual policies. This result demonstrates the strength and uniqueness of \method{} as a tactile sensor for contact-rich manipulation.

\subsubsection{Comparison across sensors}
To better contextualize the significance of this result, we present a replaceability comparison with DIGIT~\cite{lambeta2020digit} and ReSkin~\cite{bhirangi2021reskin} sensors. We collect two additional datasets of 96 demonstration trajectories each for the plug insertion task with these sensors similar to \method{}. Replaceability is evaluated by swapping the training skin for a new skin during evaluation as outlined in the previous section. Success rates from 10 evaluations across three seeds for each setting are reported in Table~\ref{tab:skin-swapping}. 

\begin{table}[htbp]
    \centering
    \caption{Success rates (out of 10) for policies when swapping out tactile skins. All statistics computed over 3 training seeds }
    \begin{tabular}{l c c c c}
         Task & Cameras only & \multicolumn{2}{c}{Cameras + Skin} \\ \cmidrule{3-4}
         & & Original skin & Swapped skin \\
         \midrule
         \multicolumn{4}{l}{\textit{Cross-instance generalization}} \\
         Plug Insertion & $1.7 \pm 0.6$ & $6.7 \pm 1.5$ & $5.3 \pm 2.5$ & \\
         Card Swiping & $2.0 \pm 1.0$ & $7.0 \pm 1.7$ & $6.3 \pm 0.6$ & \\
         USB Insertion & $1.7 \pm 1.2$ & $5.7 \pm 1.5$ & $3.0 \pm 1.0$ & \\ \midrule \midrule
         \multicolumn{4}{l}{\textit{Comparison across sensors -- Plug Insertion}} \\
         \method & $1.7 \pm 0.6$ & $\mathbf{6.7 \pm 1.5}$ & $\mathbf{5.3 \pm 2.5}$ & \\
         ReSkin & $1.7 \pm 1.2$ & $6.0 \pm 1.7$ & $1.7 \pm 1.2$ & \\
         DIGIT & $1.7 \pm 1.5$ & $2.3 \pm 0.6$ & $1.3 \pm 0.6$ \\
         \bottomrule
    \end{tabular}
    \label{tab:skin-swapping}
\end{table}

Based on these results, we find that visuotactile policies trained with ReSkin and \method{} have similar performance on solving the plug insertion task. However, when the skin is replaced, the performance of the ReSkin policy falls 43\% to the same level as the camera-only policy, while the performance of \method{} policies only drops by 13\%. This transferability is evidence of \method{}'s superior signal consistency, and is a significant boost to scaling efforts like training large tactile models as well as real world deployment of models trained in the laboratory.

\begin{figure}
    \centering
    \includegraphics[width=\linewidth]{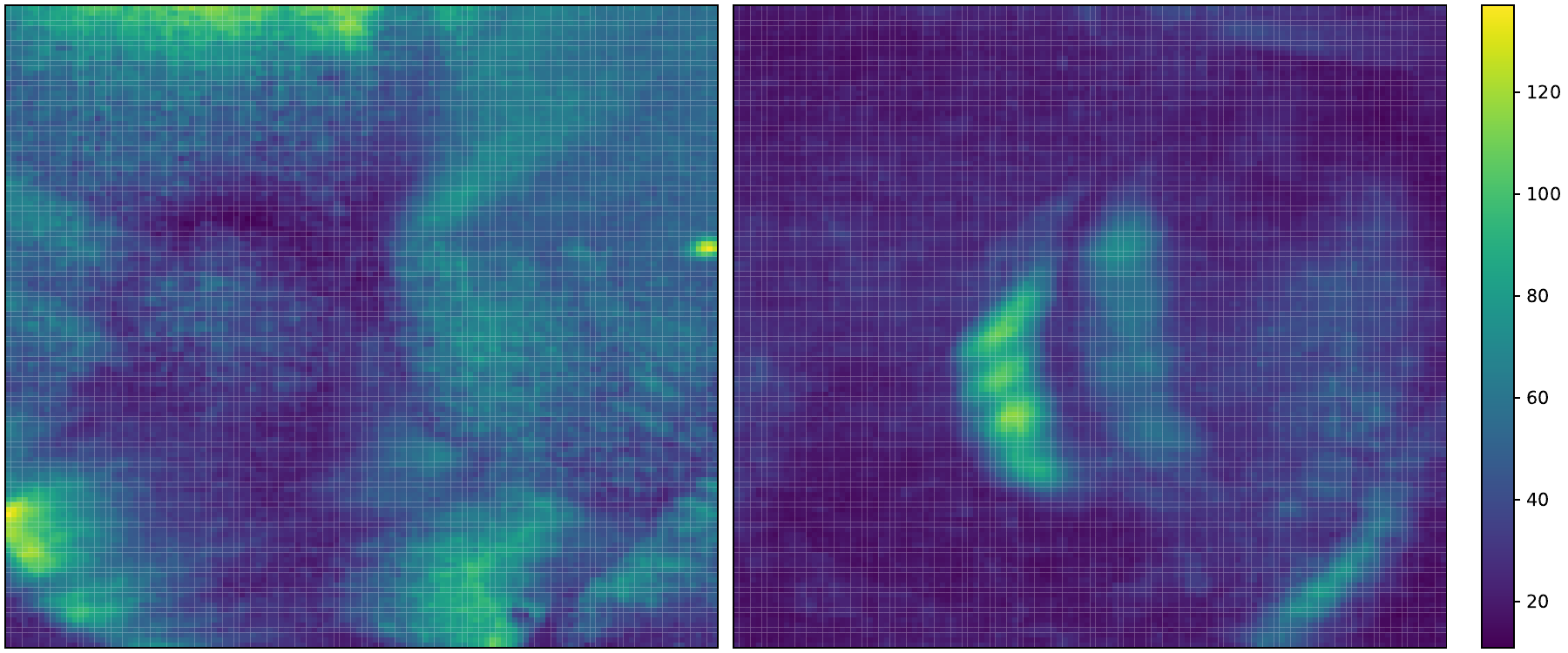}
    \caption{Pixel-wise difference between two different DIGIT sensor instances (left) and the maximum difference in response of one DIGIT sensor on the task of plug insertion across 96 demos (right).}
    \label{fig:digit-comparisons}
\end{figure}

An unexpected result from these experiments was the poor success rate of policies trained using the DIGIT sensor which has been shown to be successful in other visuotactile tasks, perhaps on less precise~\cite{qi2023general} or less reactive~\cite{suresh2023neural} ones, in prior work. Consequently, while there is still a gap in performance, we don't see a significant gap between the poor performance of the visuotactile policy on the original and swapped DIGIT skins. However, the high variability across instances of the DIGIT sensors is previously documented~\cite{suresh2023neural}, and we find that a closer look at the DIGIT signal from our plug insertion dataset indicates that even if it were possible to train more performant policies, they are unlikely to generalize across instances. Across the 96 demonstration trajectories from the plug insertion, we compute the \textit{maximum} change in pixel values across channels and across trajectories, and compare it against the pixel-wise differences between the original and a swapped instance of the DIGIT sensor in Fig.~\ref{fig:digit-comparisons}. Since the \textit{maximum} change in sensory measurement through the course of the interaction is comparable to the difference in signal between two instances, it is unlikely that policies trained on one sensor will generalize to new instances.

\section{Conclusion and Limitations}
In this paper, we present \method{}, a new magnetic tactile sensor. \method{} is versatile, self-adhering and improves on signal consistency across different instances of the skin. Furthermore, to the best of our knowledge, \method{} is the first sensor to demonstrate zero-shot generalization of visuotactile policies to new instances of the tactile skin. As we work towards developing more capable and performant models utilizing tactile data, \method{} is the first step towards ensuring that large datasets of tactile data can be collected and effectively used for training useful, generalizable models. This work opens the door to a host of exciting applications of \method{}. It could be incorporated into a large-scale data collection tool such as UMI~\cite{chi2024universal} or the Stick~\cite{shafiullah2023bringing, etukuru2024robot}. Future work could also investigate simple calibration schemes or conditional learning frameworks to completely close the gap between training and swapped skin instances. Deeper investigations into standardizing the fabrication procedure could also help further improve signal consistency.

Despite its potential, \method{} still inherits some of the drawbacks of the ReSkin sensor, primary among them being interference from magnetic and ferromagnetic objects in the environment. Using machine learning approaches for noise reduction in magnetic sensors~\cite{bhirangi2024hierarchical} or improving the skin design to incorporate a Faraday cage could help resolve these difficulties, and take tactile sensing one step closer to being a first-class citizen in robotics.

Our experiments were performed using DIGIT sensors with the standard, commercially available fingertip gel, but prior work has found some success learning visuotactile policies using optical sensors with dotted skins~\cite{dong2019tactile, dong2021tactile}. An interesting direction for future work could be comparing the performance of behavior cloning policies using different gel tips for optical sensors. However, while this may improve learning performance with a single skin, cross-instance generalizability might still require significant changes in the fabrication of optical sensors.





\section*{ACKNOWLEDGMENTS}
Special thanks for Krishna Bodduluri, Mike Lambeta, and team from Meta AI Research for providing the DIGIT sensors for comparison.
NYU authors are supported by grants from Honda, Hyundai, NSF award 2339096 and ONR awards N00014-21-1-2758 and N00014-22-1-2773. LP is supported by the Packard Fellowship.

\bibliographystyle{IEEEtran}
\bibliography{IEEEabrv, references}

\end{document}